\documentclass[runningheads]{llncs}
\usepackage{graphicx}
\usepackage{amsmath,amssymb} 
\usepackage{color}
\usepackage{booktabs}
\begin{document}
\title{NAM: Non-Adversarial Unsupervised Domain Mapping} 

\titlerunning{NAM: Non-Adversarial Unsupervised Domain Mapping}
\author{Yedid Hoshen\inst{1}\and Lior Wolf\inst{1,2}}
%
\authorrunning{Y. Hoshen and L. Wolf}

\institute{Facebook AI Research \and Tel Aviv University}
\maketitle  

\begin{abstract}
Several methods were recently proposed for the task of translating images between domains without prior knowledge in the form of correspondences. The existing methods apply adversarial learning to ensure that the distribution of the mapped source domain is indistinguishable from the target domain, which suffers from known stability issues. In addition, most methods rely heavily on ``cycle'' relationships between the domains, which enforce a one-to-one mapping. In this work, we introduce an alternative method: Non-Adversarial Mapping (NAM), which separates the task of target domain generative modeling from the cross-domain mapping task. NAM relies on a pre-trained generative model of the target domain, and aligns each source image with an image synthesized from the target domain, while jointly optimizing the domain mapping function. It has several key advantages: higher quality and resolution image translations, simpler and more stable training and reusable target models. 
Extensive experiments are presented validating the advantages of our method.

\end{abstract}

\section{Introduction}

The human ability to think in spontaneous analogies motivates the field of unsupervised domain alignment, in which image to image translation is achieved without correspondences between samples in the training set. Unsupervised domain alignment methods typically operate by finding a function for mapping images between the domains so that after mapping, the distribution of mapped source images is identical to that of the target images. 

Successful recent approaches, e.g. DTN~\cite{02200}, CycleGANs~\cite{CycleGAN2017} and DiscoGAN~\cite{discogan}, utilize Generative Adversarial Networks (GANs)~\cite{gan} to model the distributions of the two domains, $\cal X$ and $\cal Y$. GANs are very effective tools for generative modeling of images, however they suffer from instability in training, making their use challenging. The instability typically requires careful choice of hyper-parameters and often multiple initializations due to mode collapse. Current methods also make additional assumptions that can be restrictive, e.g., DTN assumes that a pre-trained high-quality domain specific feature extractor exists which is effective for both domains. This assumption is good for the domain of faces (which is the main application of DTN) but may not be valid for all cases. CycleGAN and DiscoGAN make the assumption that a transformation $T_{XY}$ can be found for every $\cal X$-domain image $x$ to a unique $\cal Y$-domain image $y$, and another transformation $T_{YX}$ exists between the $\cal Y$ domain and the original $X$-domain image $y = T_{XY}(x)$, $x=T_{YX}(y)$. This is problematic if the actual mapping is many-to-one or one-to-many, as in super-resolution or coloring.

We propose a novel approach motivated by cross-domain matching. We separate the problem of modeling the distribution of the target domain from the source to target mapping problem. We assume that the target image domain distribution is parametrized using a generative model. This model can be trained using any state-of-the-art unconditional generation method such as GAN~\cite{dcgan}, GLO~\cite{glo}, VAE~\cite{vae} or an existing graphical or simulation engine. Given the generative model, we solve an unsupervised matching problem between the input $\cal Y$ domain images and the $\cal X$ domain. For each source input image $y$, we synthesize an $\cal X $ domain image $G(z_y)$, and jointly learn the mapping function $T()$, which maps images from the $X$ domain to the $\cal Y$ domain. The synthetic images and mapping function are trained using a reconstruction loss on the input $\cal Y$ domain images.   

Our method is radically different from previous approaches and it presents the following advantages:
\begin{enumerate}
\item A generative model needs to be trained only once per target dataset, and can be used to map to this dataset from all source datasets without adversarial generative training.
\item Our method is one-way and does not assume a one-to-one relationship between the two domains, e.g., it does not use cycle-constraints. 
\item Our work directly connects between the vast literature of unconditional image generation and the task of cross-domain translation. Any progress in unconditional generation architectures can be simply plugged in with minimal changes. Specifically, we can utilize recent very high-resolution generators to obtain high quality results.
\end{enumerate}

\section{Previous Work}

\paragraph{Unsupervised domain alignment:} Mapping across similar domains without supervision has been successfully achieved by classical methods such as Congealing \cite{miller2000learning}. Unsupervised translation across very different domains has only very recently began to generate strong result, due to the advent of generative adversarial networks (GANs), and all state-of-the-art unsupervised translation  methods we are aware of employ GAN technology . As this constraint is insufficient for generating good translations, current methods are differentiated by additional constraints that they impose. 

The most popular constraint is cycle-consistency: enforcing that a sample that is mapped from $\cal X$ to $\cal Y$ and back to $\cal X$, reconstructs the original sample. This is the approach taken by DiscoGAN \cite{discogan}, CycleGAN \cite{CycleGAN2017} and DualGAN \cite{dualgan}. 
Recently, StarGAN~\cite{choi2017stargan} created multiple cycles for mapping in any direction between multiple (two or more) domains. The generator receives as input the source image as well as the specification of the target domain.

For the case of linear mappings, orthogonality has a similar effect to circularity.  
Very recently, it was used outside computer vision by several methods~\cite{zhang2017adversarial,zhang2017earth,conneau2017word,hoshen2018iterative} for solving the task of mapping words between two languages without using parallel corpora.

Another type of constraint is provided by employing a shared latent space. Given samples from two domains $\cal X$ and $\cal Y$, CoGAN~\cite{cogan}, learns a mapping from a random input vector $z$ to matching samples, one in each domain. The domains $\cal X$ and $\cal Y$ are assumed to be similar and their generators (and GAN discriminators) share many of the layers' weights, similar to~\cite{ilya}. Specifically, the earlier generator layers are shared while the top layer are domain specific. CoGAN can be modified to perform domain translation in the following way: given a sample $x \in \cal X$, a latent vector $z_x$ is fitted to minimize the distance between the image generated by the first generator $G_{\cal X}(z_x)$ and the input image $x$. Then, the analogous image in $\cal Y$ is given by $G_{\cal Y}(z_x)$. This method was shown in~\cite{CycleGAN2017} to be less effective than cycle-consistency based methods.

UNIT~\cite{liu2017unsupervised} employs an encoder-decoder pair per each domain. The latent spaces  of the two are assumed to be shared, and similarly to CoGAN, the layers that are distant from the image (the top layers of the encoder and the bottom layers of the encoder) are shared between the two domains. Cycle-consistency is added as well, and structure is added to the latent space using variational autoencoder~\cite{kingma2014auto} loss terms.

As mentioned above our method does not use adversarial or cycle-consistency constraints.

 \paragraph{Mapping using Domain Specific Features} Using domain specific features has been found by DTN~\cite{02200} to be important for some tasks. It assumed that a feature extractor can be found, for which the source and target would give the same activation values. Specifically it uses face specific features to map faces to emojis. While for some of the tasks, our work does use a ``perceptual loss'' that employs a pretrained imagenet-trained network, this is a generic feature extraction method that is not domain specific. We claim therefore that our method still qualifies as unsupervised. For most of the tasks presented, the VGG loss alone, would not be sufficient to recover good mappings between the two domains, as shown in ANGAN~\cite{hoshen2018identifying}. 

\paragraph{Unconditional Generative Modeling:}  Many methods were proposed for generative models of image distributions. Currently the most popular approaches rely on GANs and VAEs~\cite{vae}. GAN-based methods are plagued by instability during training. Many methods were proposed to address this issue for unconditional generation, e.g.,~\cite{arjovsky2017wasserstein,gulrajani2017improved,miyato2018spectral}. The modifications are typically not employed in cross-domain mapping works. Our method trains a generative model (typically a GAN), in the $\cal X$ domain separately from any $\cal Y$ domain considerations, and can directly benefit from the latest advancements in the unconditional image generation literature. GLO~\cite{bojanowski2017optimizing} is an alternative to GAN, which iteratively fits per-image latent vectors (starting from random ``noise'') and learns a mapping $G()$ between the noise vectors and the training images. GLO is trained using a reconstruction loss, minimizing the difference between the training images and those generated from the noise vectors. Differently from our approach is tackles unconditional generation rather than domain mapping. 

\section{Unsupervised Image Mapping without GANs}
\label{sec:mapping_no_gan}

In this section, we present our method - NAM -  for unsupervised domain mapping. The task we aim to solve, is finding analogous images across domains. Let $\cal X$ and $\cal Y$ be two image domains, each with some unique characteristics. For each domain we are given a set of example images. The objective is to find for every image \textit{y} in the $\cal Y$ domain, an analogous image \textit{x} which appears to come from the $\cal X$ domain but preserves the unique content of the original \textit{y} image.

\subsection{Non-Adversarial Exact Matching}
\label{subsec:namatch}

To motivate our approach, we first consider the simpler case, where we have two image domains $\cal X$ and $\cal Y$, consisting of sets of images $\{x_i\}$ and $\{y_i\}$ respectively. We assume that the two sets are approximately related by a transformation $T$, and that a matching paired image $x$ exists for every image \textit{y} in domain $\cal Y$ such that $T(x) = y$. The task of matching becomes a combination of two tasks: i) inferring the transformation between the two domains ii) finding matching pairs across the two domains. Formally this becomes:
\begin{equation}
\label{eq:angan}
L = \sum_{i}{ \|T(\sum_j M_{ij} x_j) , y_i\|}
\end{equation}
Where $M_{ij}$ is the matching matrix containing $M_{i,j}=1$ if $x_j$ and $y_i$ are matching and $0$ otherwise. The optimization is over both the transformation $T()$ as well as binary match matrix $M$.

Since the optimization of this problem is hard, a relaxation method - \textit{ANGAN} - was recently proposed~\cite{hoshen2018identifying}. The binary constraint on the matrix was replaced by the requirement that $M_{ij} \geq 0$ and $\sum_j{M_{ij}} = 1$. As optimization progresses, a barrier constraint on $M$, pushes the values of $M$ to $0$ or $1$.

ANGAN was shown to be successful in cases where exact matches exist and $T()$ is initialized with a reasonably good solution obtained by CycleGAN.

\subsection{Non-Adversarial Inexact Matching}
\label{subsec:nainmatch}

In Sec.~\ref{subsec:namatch}, we described the scenario in which exact matches exist between the images in domains $\cal X$ and $\cal Y$. In most situations, exact matches do not exist between the two domains. In such situations it is not sufficient to merely find an image $x$ in the domain $\cal X$ training set such that for a target $\cal Y$ domain image $y$, we have $y = T(x)$ as we cannot hope that such a match will exist. Instead, we need to synthesize an image $\tilde{x}$ that comes from the $\cal X$ domain distribution, and satisfies $y = T(\tilde{x})$. This can be achieved by removing the stochasticity requirement in Eq.~\ref{eq:angan}. Effectively, this models the images in the $\cal X$ domain as:
\begin{equation}
\label{eq:simplex}
\tilde{x} = \sum_j \alpha_i x_i
\end{equation}
This solution is unsatisfactory on several counts: (i) the simplex model for the $\cal X$ domain cannot hope to achieve high quality image synthesis for general images (ii) The complexity scales quadratically with the number of training images making both training and evaluation very slow.  

\begin{figure}[t]
  \centering
      
\includegraphics[width=1.0\linewidth]{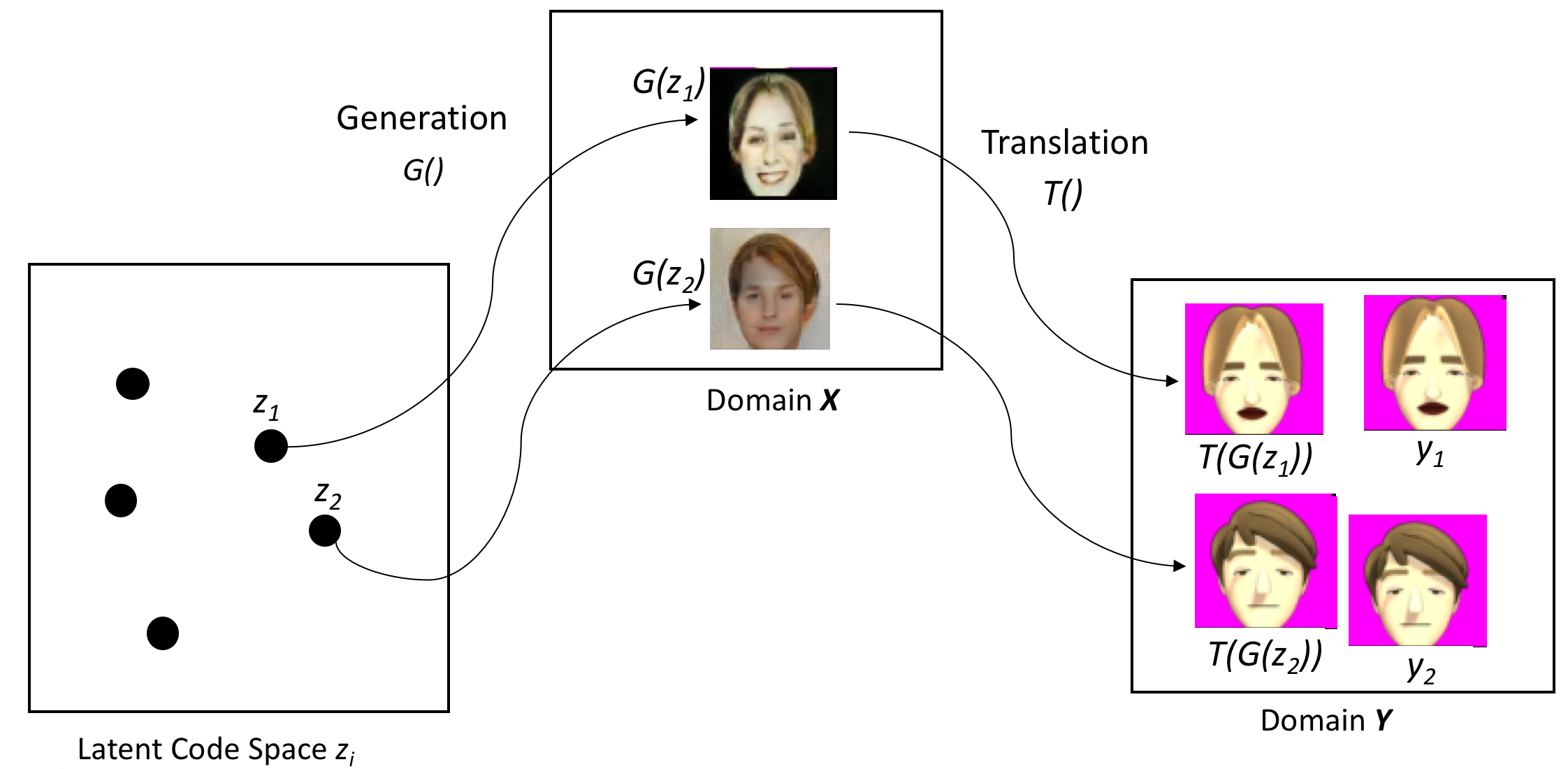}

  \caption{Given a generator $G$ for domain $\cal X$ and training samples $\{y_i\}$ in domain $\cal Y$, NAM jointly learns the transformation $T:\cal X \rightarrow \cal Y$ and the latent vectors $\{z_i\}$ that give rise to samples $\{T(G(z_i))\}$ that resemble the training images in $\cal Y$}
  \label{fig:nam_method}

\end{figure}

\subsection{Non-Adversarial Mapping (NAM)}
\label{subsec:nam}

In this section we generalize the ideas presented in the previous sections into an effective method for mapping between domains without supervision or the use of adversarial methods. 

In Sec.~\ref{subsec:nainmatch}, we showed that to find analogies between domains $\cal X$ and $\cal Y$, the method requires two components: (i) a model for the distribution of the $\cal X$ domain, and (ii) a mapping function $T()$ between domains $\cal X$ and $\cal Y$.

Instead of the linear simplex model of Sec.~\ref{subsec:nainmatch}, we propose to model the $\cal X$ domain  distribution  by a neural generative model $G(z)$, where $z$ is a latent vector. The requirements on the generative model $G()$ are such that for every image $x$ in the $\cal X$ domain distribution we can find $z$ such $x = G(z)$ and that $G()$ is compact, that is, for no $z$, will $G(z)$ lie outside the $\cal X$ domain. The task of learning such generative models, is the research focus of several communities. In this work we do not aim to contribute to the methodology of unsupervised generative modeling, but rather use the state-of-the-art modeling techniques obtained by previous approaches, for our generator $G()$. Methods which can be used to obtain generative model $G()$ include: GLO~\cite{glo}, VAE~\cite{vae}, GAN~\cite{gan} or a hand designed simulator (see for example \cite{wolf2017unsupervised}). In our method, the task of single domain generative modeling is entirely decoupled from the task of cross-domain mapping, which is the one we set to solve. 

Armed with a much better model for the $\cal X$ domain distribution, we can now make progress on finding synthetic analogies between $\cal X$ and $\cal Y$. Our task is to find for every $\cal Y$ domain image $y$, a synthetic $\cal X$ domain image $G(z_y)$ so that when mapped to the $\cal Y$ domain $y = T(G(z_y))$. The task is therefore twofold: (i) for each $y$, we need to find the latent vector $z_y$ which will synthesize the analogous $\cal X$ domain image, and (ii) the mapping function $T()$ needs to be learned.

The model can therefore be formulated as an optimization problem, where the objective is to minimize the reconstruction cost of the training images of the $\cal Y$ domain. The optimization is over the latent codes, a unique latent code $z_y$ vector for every input $\cal Y$ domain image $y$, as well as the mapping function $T()$. It is formally written as below:  
\begin{equation}
\label{eq:num1}
argmin_{T, z_y} \sum_{y \in B} \|T(G(z_y)), y\| 
\end{equation}

The model is fully differentiable, as both the generative model $G()$ and the mapping function $T()$ are parameterized by neural networks. The above objective is jointly optimized for $z_y$ and $T()$, but not for $G()$ which is kept fixed. The method is illustrated in Fig.~\ref{fig:nam_method}. 

\subsection{Perceptual Loss}
\label{subsec:perceptual}

Although the optimization described in Sec.~\ref{subsec:nam} can achieve good solutions, we found that introducing a perceptual loss, can significantly help further improve the quality of analogies. Let $\phi_i()$ be the features extracted from a deep-network at the end of the i'th block (we use VGG \cite{simonyan2014very}). The perceptual loss is given by:
\begin{equation}
\label{eq:perceptual}
\|.,.\|_{VGG}  = \sum_i \|\phi_i(T(G(z_y))), \phi_i(y)\|_1 + \|T(G(z_y)), y\|_1
\end{equation}

The final optimization problem becomes:
\begin{equation}
\label{eq:nam_final}
argmin_{T, z_y} \sum_{y \in B} \|T(G(z_y)), y\|_{VGG}
\end{equation}

The VGG perceptual loss was found by several recent papers \cite{chen2017photographic,zhang2018unreasonable} to give perceptually pleasing results. There have been informal claims in the past that methods using perceptual loss functions should count as supervised. We claim that the perceptual loss does not make our method supervised, as the VGG network does not come from our domains and does not require any new labeling effort. Our view is that taking advantage of modern feature extractors will benefit the field of unsupervised learning in general and unsupervised analogies in particular.

\subsection{Inference and Multiple Solutions}
\label{subsec:inference}

Once training has completed, we are now in possession of the mapping function $T()$ which is now fixed (the pre-trained $G()$ was never modified as a part of training). 

To infer the analogy of a new $\cal Y$ domain image $y$, we need to recover the latent code $z_y$ which would yield the optimal reconstruction. The mapping function $T()$ is now fixed, and is not modified after training. Inference is therefore performed via the following optimization:
\begin{equation}
\label{eq:nam_inference}
argmin_{z_y} \|T(G(z_y)), y\|
\end{equation}

The synthetic $\cal X$ domain image $G(z_y)$ is our proposed solution to $\cal Y$ domain image $y$.

This inference procedure is a non-convex optimization problem. Different initializations, yield different final analogies. Let us denote initialization $z_0^t$ where $t$ is the ID of the solution. At the end of the optimization procedure for each initialization, the synthetic images $G(z^t)$ yield multiple proposed analogies for the task. We find $G(z^0)...G(z^T)$ are very diverse when in fact many analogies are available. For example, when the $\cal X$ domain is Shoes and the $\cal Y$ domain is Edges, there are many shoes that can result in the same edge image.    

\subsection{Implementation Details}

In this section we give a detailed description of the procedure used to generate the experiments presented in this paper.

\textit{$\cal X$ domain generative model $G(.)$:} Our method takes as input a pre-trained generative model for the $\cal X$ domain. In our MNIST, SVHN and cars, Edges2 (Shoes,Handbags) experiments, we used DCGAN \cite{dcgan} with (32,32,32,100,100) latent dimensions. The low resolution face image generator was trained on celebA and used the training method of~\cite{miyato2018spectral}. The high resolution face generator is provided by~\cite{karras2017progressive} and the Dog generator by~\cite{Han17stackgan2}. The hyperparameters of all trained generators were set to their default value. In our experiments GAN unconditional generators provided more compelling results than competing SOTA methods such as GLO and VAE. 

\textit{Mapping function $T(.)$:} The mapping function was designed so that it is powerful enough but not too large as to overfit. Additionally, it needs to preserve image locality, in the case of spatially aligned domains. We elected to use a network with an architecture based on~\cite{chen2017photographic}. We found that as we only rely on the networks to find correspondences rather than generate high-fidelity visual outputs, small networks were the preferred choice. We used a similar architecture to \cite{chen2017photographic}, with a single layer per scale, and linearly decaying number of filters per layer starting with 4$F$, and decreasing by $F$ with every layer. $F=8$ for SVHN and MNIST and $F=32$ for the other experiments. 

\textit{Optimization:} We optimized using SGD with ADAM~\cite{adam}. For all datasets we used a learning rate of $0.03$ for the latent codes $z_y$ and $0.001$ for the mapping function $T(.)$ (due to the uneven update rates of each $z_y$ and $T()$). On all datasets training was performed on $2000$ randomly selected examples (a subset) from the $\cal Y$ domain. Larger training sets were not more helpful as each $z_y$ is updated less frequently.

\textit{Generating results:} The $\cal X$ domain translation of $\cal Y$ domain image $y$ is given by $G(z_y)$, where $z_y$ is the latent code found in optimization. The $\cal X \to \cal Y$ mapping $T(x)$, typically resulted in weaker results due to the relatively shallow architecture selected for $T(.)$. A strong $T(.)$ can be trained by calculating a set of $G(z_y)$ and $y$ (obtained using NAM), and training a fully-supervised network $T(.)$, e.g. as described by \cite{chen2017photographic}. A similar procedure was carried out in \cite{hoshen2018identifying}.

\section{Experiments}
\label{sec:exp}

To evaluate the merits of our method, we carried out an extensive set of qualitative and quantitative experiments. 

\begin{figure}[t]
  \centering

    \begin{tabular}{ccc}
&    {SVHN$\rightarrow$MNIST} & {MNIST$\rightarrow$SVHN}\\
\includegraphics[width=0.08810\linewidth]{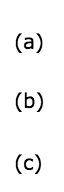} &\includegraphics[width=0.45\linewidth]{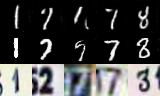}&
\includegraphics[width=0.45\linewidth]{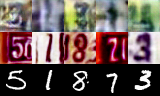} \\

    \end{tabular}

  \caption{Converting digits between SVHN and MNIST (both directions). (a) CycleGAN results (b) NAM results (c) the input images.}
  \label{fig:mnist_svhn}

\end{figure}

\textbf{SVHN-MNIST Translation:}   We evaluated our method on the SVHN-MNIST translation task. Although SVHN \cite{svhn} and MNIST \cite{mnist} are simple datasets, the mapping task is not trivial. The MNIST dataset consists of simple handwritten single digits written on black background. In contrast, SVHN images are taken from house numbers and typically contain not only the digit of interest but also parts of the adjacent digits, which are nuisance information. We translate in both directions SVHN$\to$MNIST and MNIST$\to$SVHN. The results are presented in Fig.~\ref{fig:mnist_svhn}. We can observe that in the easier direction of SVHN$\to$MNIST, in which there is information loss, NAM resulted in more accurate translations than CycleGAN. In the reverse direction of MNIST$\to$SVHN, which is harder due to information gain, CycleGAN did much worse, whereas NAM was often successful. Note that perceptual loss was not used in the MNIST$\to$SVHN translation task.       

\begin{table}[t]
  \centering
      
  \caption{Translation quality measured by translated digit classification accuracy (\%)}
  \label{tab:svhn_mnist_acc}

    \begin{tabular}{lcc}
    \toprule
	 & \textit{SVHN$\to$MNIST} & \textit{MNIST$\to$SVHN}   \\    
    \midrule
    \textit{CycleGAN} & 26.8 & 17.7\\
    \textit{NAM} & 33.3 & 31.9\\
	 \bottomrule
    \end{tabular}
\end{table}

We performed a quantitative evaluation of the quality of SVHN$\leftrightarrow$MNIST translation. This was achieved by mapping an image from the one dataset to appear like the other dataset, and classifying it using a pre-trained classifier trained on the clean target data (the classifier followed a NIN architecture~\cite{nin}, and achieved test accuracies of around 99.5\% on MNIST and 95.0\% on SVHN). The results are presented in Tab.~\ref{tab:svhn_mnist_acc}. We can see that the superior translations of NAM are manifested in higher classification accuracies. 

\textbf{Edges2Shoes:} The task of mapping edges to shoes is commonly used to qualitatively evaluate unsupervised domain mapping methods. The two domains are a set of Shoe images first collected by \cite{edges2shoes}, and their edge maps. The transformation between an edge map and the original photo-realistic shoe image is non-trivial, as much information needs to be hallucinated.

\begin{figure}
  \centering
    \begin{tabular}{cc}
\includegraphics[width=0.45\linewidth]{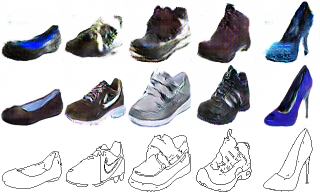} &
\includegraphics[width=0.45\linewidth]{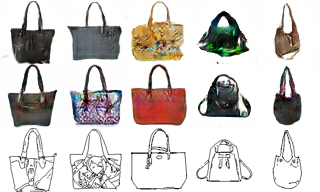} \\
 (a) & (b) \\
 \includegraphics[width=0.45\linewidth]{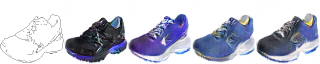} &
\includegraphics[width=0.45\linewidth]{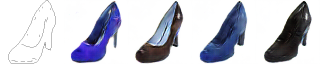}\\
 (c) & (d) \\

    \end{tabular}
      \caption{(a) Comparison of NAM and DiscoGAN for Edges$2$Shoes. Each triplet shows NAM (center row) vs. DiscoGAN (top row) for a given input (bottom row). (b) A similar visualization for Edges$2$Handbags. (c,d) NAM mapping from a single source edge image (shown first) for different random initializations.}
  \label{fig:shoes_bags}

\end{figure}

Examples of NAM and DiscoGAN results can be seen in Fig.~\ref{fig:shoes_bags}(a). The higher quality of the analogies generated by NAM is apparent. This stems from using a pre-learned generative model rather than learning jointly with mapping, which is hard and results in worse performance. We also see the translations result in more faithful analogies. Another advantage of our method is the ability to map one input into many proposed solutions. Two examples are shown in Fig.~\ref{fig:shoes_bags}(c) and (d). It is apparent that the solutions all give correct analogies, however they give different possibilities for the correct analogy. This captures the one-to-many property of the edge to shoes transformation.

\begin{figure}[t]
\resizebox{\textwidth}{!}{%
    \begin{tabular}{c|c}
\textit{Target VAE-h VAE-l GLO DCGAN SNGAN}&
\textit{Target VAE-h VAE-l GLO DCGAN SNGAN}\\
    \midrule
\includegraphics[width=0.45\linewidth]{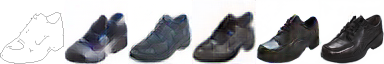} &
\includegraphics[width=0.45\linewidth]{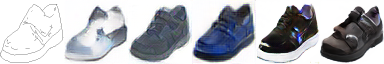} \\
\includegraphics[width=0.45\linewidth]{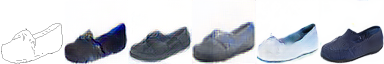} &
\includegraphics[width=0.45\linewidth]{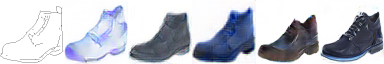} \\
\includegraphics[width=0.45\linewidth]{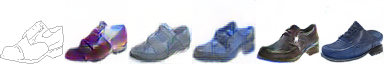}&
\includegraphics[width=0.45\linewidth]{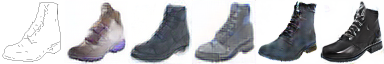} \\
    \end{tabular}
    }
    \caption{Comparison of NAM results for different generators}
    \label{fig:several}
\end{figure}

As mentioned in the method description, NAM requires high-quality generators, and performs better for better pre-trained generators. In Fig.~\ref{fig:several} we show NAM results for generators trained with: VAE \cite{vae} with high (VAE-h) and low (VAE-l) regularization, GLO \cite{glo}, DCGAN \cite{dcgan} and Spectral-Normalization GAN \cite{miyato2018spectral}. We can see from the results that NAM works is all cases. however results are much better for the best generators (DCGAN, Spectral-Norm GAN).

\textbf{Edges2Handbags:} The Edges2Handbags \cite{edges2bags} dataset is constructed similarly to Edges2Shoes. Sample results on this dataset can be seen in Fig.~\ref{fig:shoes_bags}(b). The conclusions are similar to Edges2Shoes: NAM generates analogies that are both more appealing and more precise than DiscoGAN.

\textbf{Shoes2Handbags:} One of the major capabilities displayed by DiscoGAN is being able to relate domains that are very different. The example shown in~\cite{discogan}, of mapping images of handbags to images of shoes that are semantically related, illustrates the ability of making distant analogies.

\begin{figure}[t]
  \centering
      
\includegraphics[width=1.0\linewidth]{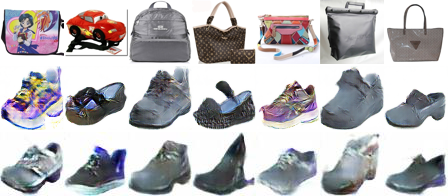}

  \caption{Example results for mapping from bags (original images - top) to shoes. NAM  mapped images (center) are clearly better than DiscoGAN mapped images (bottom).}
  \label{fig:bags2shoes}

\end{figure}

In this experiment we show that NAM is able to make analogies between handbags and shoes, resulting in higher quality solutions than those obtained by DiscoGAN. In order to achieve this, we replace the reconstruction VGG loss by a Gram matrix VGG loss, as used in Style Transfer \cite{styletransfer}. DiscoGAN also uses a Gram matrix loss (with feature extracted from its discriminator). For this task, we also add a skip connection from $G(z)$, as the domains are already similar under a style loss. 

Example images can be seen in Fig.~\ref{fig:bags2shoes}. The superior quality of the NAM generated mapped images is apparent. The better quality is a result of using an interpretable and well understood non-adversarial loss which is quite straight forward to optimize. Another advantage comes from being able to "plug-in" a high-quality generative model.

\begin{table}[t]
  \centering
      
  \caption{Car2Car root median residual deviation from linear alignment (lower is better). }
  \label{tab:car2car}

    \begin{tabular}{cc}
    \toprule
	 \textit{DiscoGAN} & \textit{NAM}   \\    
     \midrule
    13.81 & 1.47\\
	 \bottomrule
    \end{tabular}
\end{table}

\begin{figure}[t]
  \centering
      
\includegraphics[width=1.0\linewidth]{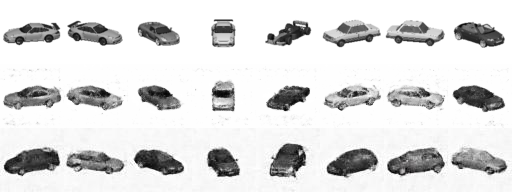}

  \caption{Example results for mapping across two sets of car models at different orientations. Although DiscoGAN (bottom) does indeed preserve orientation of the original images (top) to some extent, NAM (center) preserves both orientation and general car properties very accurately - despite the target domain containing few sports cars. }
  \label{fig:car2car}

\end{figure}

\textbf{Car2Car:} The Car2Car dataset is a standard numerical baseline for cross-domain image mapping. Each domain consists of a set of different cars, presented in angles varying from -75 tp 75 degrees. The objective is to align the two domains such that a simple relationship exists between orientation of car image $y$ and mapped image $x$ (typically, either the orientation of $x$ and $y$ should be equal or reversed). A few cars mapped by NAM and DiscoGAN can be seen in Fig.~\ref{fig:car2car}. Our method results in a much cleaner mapping. We also quantitatively evaluate the mapping, by training a simple regressor on the car orientation in the $\cal X$ domain, and comparing the ground-truth orientation of $y$ with the predicted orientation of the mapped image $x$. We evaluate using the root median residuals (as the regressor sometimes flips orientations of -75 to 75 resulting in anomalies). For car2car, we used a skip connection from $G(z)$ to the output. Results are seen in Tab.~\ref{tab:car2car}. Our method significantly outperforms DiscoGAN. Interestingly, on this task, on this task, it was not necessary to use a perceptual loss, a simple Euclidean pixel loss was sufficient for a very high-quality solution on this task. As a negative result, on the car2head task i.e. mapping between car images and images of heads of different people at different azimuth angles; NAM did not generate a simple relation between the orientations of the cars and heads but a more complex relationship. Our interpretation from looking at results is that black cars were inversely correlated with the head orientation, whereas white cars were positively correlated. 

\begin{figure}[t]
  \centering
      
\includegraphics[width=1.0\linewidth]{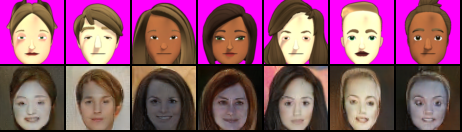}

  \caption{Example results for mapping Avatars (top) to Faces (bottom) using NAM. }
  \label{fig:emoji_lr}

\end{figure}

\textbf{Avatar2Face:} One of the first applications of cross-domain translation was face to avatar generation by DTN \cite{02200}. This was achieved by using state-of-the-art face features, and ensured the features are preserved in the original face and the output avatar ($f$-constancy). Famously however, DTN does not generate good results on avatar2face generation, which involves adding rather than taking away information. Due to the many-to-one nature of our approach, NAM is better suited for this task. In fig.~\ref{fig:emoji_lr} we present example images of our avatar2face conversions. This was generated by a small generative model with a DCGAN \cite{dcgan} architecture, trained using Spectral Normalization GAN \cite{miyato2018spectral} using celebA face images. The avatar dataset was obtained from the authors of~\cite{wolf2017unsupervised}.  

\textbf{Plugging in State-of-the-Art Generative models:} One of the advantages of our method is the independence between mapping and generative modeling. The practical consequence is that any generative model, even very large models that take weeks to train, can be effortlessly plugged into our framework. We can then map any suitable source domain to it, very quickly and efficiently.

Amazing recent progress has been recently carried out on generative modeling. One of the most striking examples of it is Progressive Growing of GANs (PGGAN) \cite{karras2017progressive}, which  has yielded generative models of faces with unprecedented resolutions of 1024X1024. The generative model training took 4 days of 8 GPUs, and the architecture selection is highly non-trivial. Including the training of such generative models in unsupervised domain mapping networks is therefore very hard.

For NAM, however, we simply set $G()$ as the trained generative model from the authors' code release. A spatial transformer layer, with parameters optimized by SGD per-image, reduced the model outputs to the Avatar scale (which we chose to be $64X64$). We present visual results in Fig.~\ref{fig:emoji_hr}. Our method is able to find very compelling analogous high-resolution faces. Scaling up to such high resolution would be highly-nontrivial with state-of-the-art domain translation methods. We mention that DTN \cite{02200}, the state-of-the-art approach for unsupervised face-to-emoji mapping, has not been successful at this task, even though it uses domain specific facial features.

\begin{figure}[t]
\centering
\begin{tabular}{cc}
\includegraphics[width=0.1\linewidth]{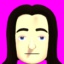}&\includegraphics[width=0.9\linewidth]{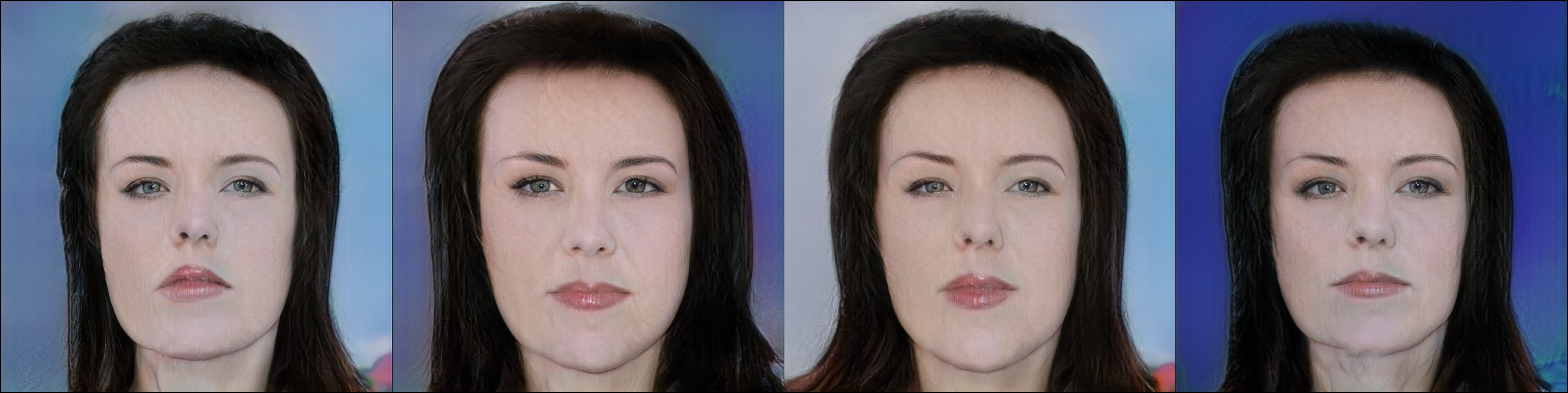}\\
\includegraphics[width=0.1\linewidth]{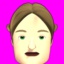}&\includegraphics[width=0.9\linewidth]{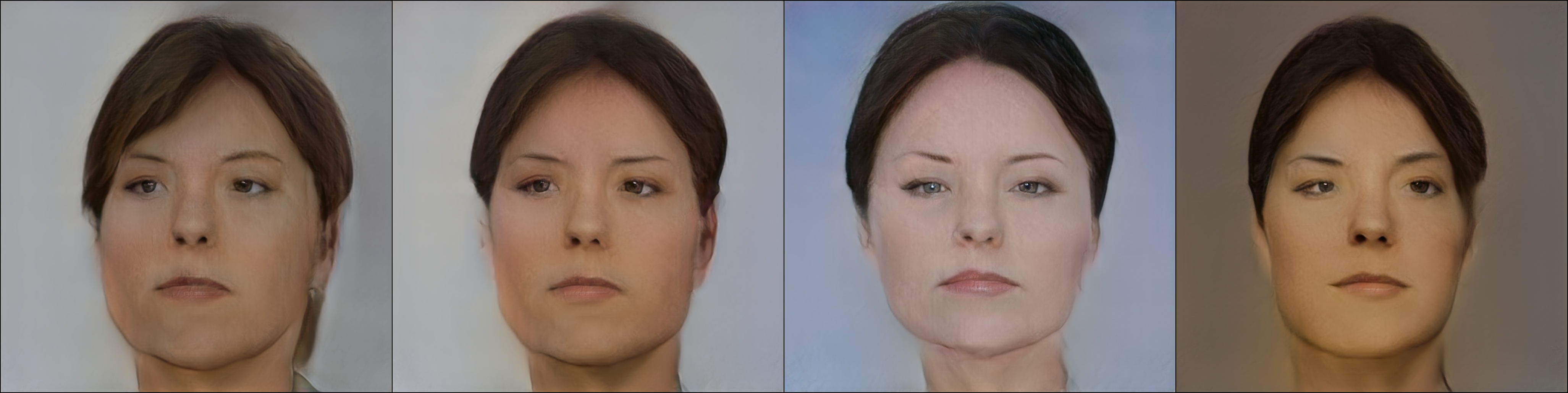}\\
\includegraphics[width=0.1\linewidth]{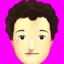}&\includegraphics[width=0.9\linewidth]{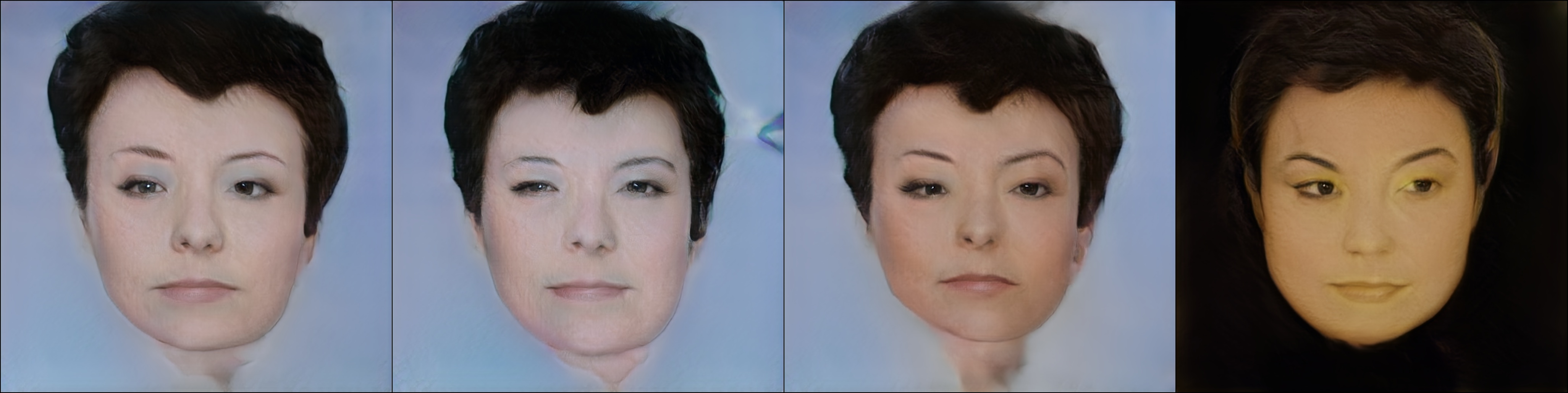}\\
(Emoji) & (Mapped faces)\\
\end{tabular}
\caption{One-to-many high-resolution mapping from Avatars to Faces using the pre-trained generator from \cite{karras2017progressive}}
\label{fig:emoji_hr}
\end{figure}

To show the generality of our approach, we also mapped Avatars to Dog images. The generator was trained using StackGAN-v2 \cite{Han17stackgan2}. We plugged in the trained generators from the publicly released code into NAM. Although emoji to dogs is significantly more distant than emoji to human face (all the Avatars used, were human faces), NAM was still able to find compelling analogies.

\begin{figure}[t]
\centering
\begin{tabular}{cccccc}
\includegraphics[width=0.1\linewidth]{progressive/avatar14.jpg}&\includegraphics[width=0.16\linewidth]{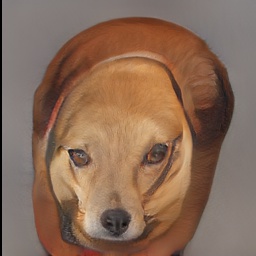} & \includegraphics[width=0.1\linewidth]{progressive/avatar2.jpg} & \includegraphics[width=0.16\linewidth]{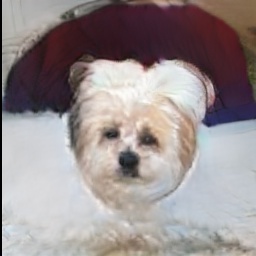} & \includegraphics[width=0.1\linewidth]{progressive/avatar15.jpg} & \includegraphics[width=0.16\linewidth]{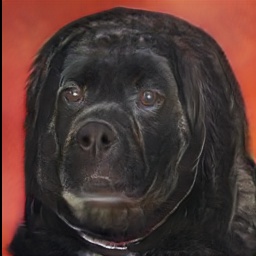}\\
(Emoji1) & (Mapped1) & (Emoji2) & (Mapped2) & (Emoji3) & (Mapped3)\\
\end{tabular}
\caption{High-resolution mapping from Avatars to Dogs, using the pre-trained generator from \cite{Han17stackgan2}.}
\label{fig:emoji_hr}
\end{figure}

\section{Discussion}

Human knowledge acquisition typically combines existing knowledge with new knowledge obtained from novel domains. This process is called blending~\cite{fauconnier2002wwt}. Our work (as most of the existing literature) focuses on the mapping process i.e. being able to relate the information from both domains, but does not deal with the actual blending of knowledge. We believe that blending, i.e., borrowing from both domains to create a unified view that is richer than both sources would be an extremely potent direction for future research.

An attractive property of our model, is the separation between the acquisition of the existing knowledge and the fitting of a new domain. The preexisting knowledge is modeled as the generative model of domain $\cal X$, given by $G$; The fitting process includes the optimization of a learned mapper from domain $\cal X$ to domain $\cal Y$, as well as identifying exemplar analogies $G(z_y)$ and $y$. 

A peculiar feature of our architecture, is that function $T()$ maps from the target ($\cal X$ domain) to the source ($\cal Y$ domain) and not the other way around. Mapping in the other direction would fail, since it can lead to a form of mode-collapse, in which all $\cal Y$ samples are mapped to the same generated $G(z)$ for a fixed $z$. While additional loss terms and other techniques can be added in order to avoid this, mode collapse is a challenge in generative systems and it is better to avoid the possibility of it altogether. Mapping as we do avoids this issue.

\section{Conclusions}
\label{sec:conc}

Unsupervised mapping between domains is an exciting technology with many applications. While existing work is currently dominated by adversarial training, and relies on cycle constraints, we present results that support other forms of training.

Since our method is very different from the existing methods in the literature, we have been able to achieve success on tasks that do not fit well into other models. Particularly, we have been able to map low resolution face avatar images into very high resolution images. On lower resolution benchmarks, we have been able to achieve more visually appealing and quantitatively accurate analogies. 

Our method relies on having a high quality pre-trained unsupervised generative model for the $\cal X$ domain. We have shown that we can take advantage of very high resolution generative models, e.g.,  \cite{karras2017progressive,Han17stackgan2}. As the field of unconditional generative modeling progresses, so will the quality and scope of NAM.    

\bibliographystyle{splncs04}
\bibliography{gans}

\end{document}